\documentclass{article}
\usepackage{spconf,amsmath,graphicx,hyperref,booktabs}

\makeatletter
\let\oldthebibliography\thebibliography
\def\thebibliography#1{%
  \oldthebibliography{#1}%
  \fontsize{9pt}{10pt}\selectfont    
  \setlength{\itemsep}{0pt plus 0.2pt}%
  \setlength{\parskip}{0pt}%
  \setlength{\parsep}{0pt}%
}
\makeatother

\title{SHARED LATENT REPRESENTATION FOR JOINT TEXT-TO-AUDIO-VISUAL SYNTHESIS}
\name{\parbox{\linewidth}{\centering Dogucan Yaman\textsuperscript{1,*}\thanks{*The authors contributed equally.}  \qquad Seymanur Akti\textsuperscript{1,*} \qquad Fevziye Irem Eyiokur\textsuperscript{1}\qquad Alexander Waibel\textsuperscript{1,2} }}

\address{\textsuperscript{1}Karlsruhe Institute of Technology, \textsuperscript{2}Carnegie Mellon University}
\begin{document}
%
\maketitle

\begin{abstract}
We propose a text-to-talking-face synthesis framework leveraging latent speech representations from HierSpeech++. A Text-to-Vec module generates Wav2Vec2 embeddings from text, which jointly condition speech and face generation. To handle distribution shifts between clean and TTS-predicted features, we adopt a two-stage training: pretraining on Wav2Vec2 embeddings and finetuning on TTS outputs. This enables tight audio-visual alignment, preserves speaker identity, and produces natural, expressive speech and synchronized facial motion without ground-truth audio at inference. Experiments show that conditioning on TTS-predicted latent features outperforms cascaded pipelines, improving both lip-sync and visual realism.
\end{abstract}

\section{Introduction}
\label{sec:intro}

Generating realistic talking-face videos directly from text, while simultaneously producing high-quality speech, remains a challenging task for applications such as virtual avatars, face-dubbing~\cite{waibel2023face}, and digital assistants. Traditional talking face generation (TFG) models trained on ground-truth audio often suffer from temporal misalignment and reduced performance when exposed to synthetic speech. Some existing pipelines adopt a cascaded approach, where text is first converted to speech and then the audio drives facial animation~\cite{wang2023text, zhang2022text2video, ye2023ada}. While effective to some extent, these methods are prone to domain shift and error accumulation, as the talking-face model is not trained on TTS-generated audio.

Other approaches mitigate this issue by creating shared latent representations for text and audio~\cite{mitsui2023uniflg, choi2024text} or by using feature fusion techniques to incorporate text-enriched features into TFG~\cite{diao2025ft2tf}. More recent works employ advanced generative models to jointly synthesize speech and talking faces~\cite{jang2024faces, wang2025omnitalker}, demonstrating the benefits of unified modeling for audio-visual coherence.

In our approach, we adopt a simple yet effective adaptation strategy. We leverage a Text-to-Vec (TTV) module to generate intermediate latent speech features directly from text. These features serve as a shared representation for both speech reconstruction and talking-face generation, ensuring tight audio-visual alignment. We then adapt the talking face generator to these intermediate TTS-predicted features, addressing the domain shift that occurs between clean, pretrained audio features and TTS outputs. By conditioning the generator on these predicted representations, we avoid the limitations of cascaded pipelines and enable existing TFG models to handle synthetic speech more effectively, improving both lip–speech synchronization and overall realism.
Our contributions are as follows:
(1) To the best of our knowledge, we present the first joint text-to-audio-visual synthesis for face dubbing.
(2) We propose a two-stage training strategy for talking face generation that learns a shared latent space and adapts effectively to TTS-predicted features.
(3) We conduct extensive experiments demonstrating that our method achieves competitive performance while enabling direct text-to-audio-video generation.
This is particularly important since generating audio and video in parallel from a joint space is crucial as it guarantees natural audio-lip synchronization and coherent audio-visual alignment.
It also eliminates the need for a cascaded system.

\section{METHOD}
\label{sec:format}

\begin{figure*}[ht!]
    \centering
    \includegraphics[width=0.8\linewidth]{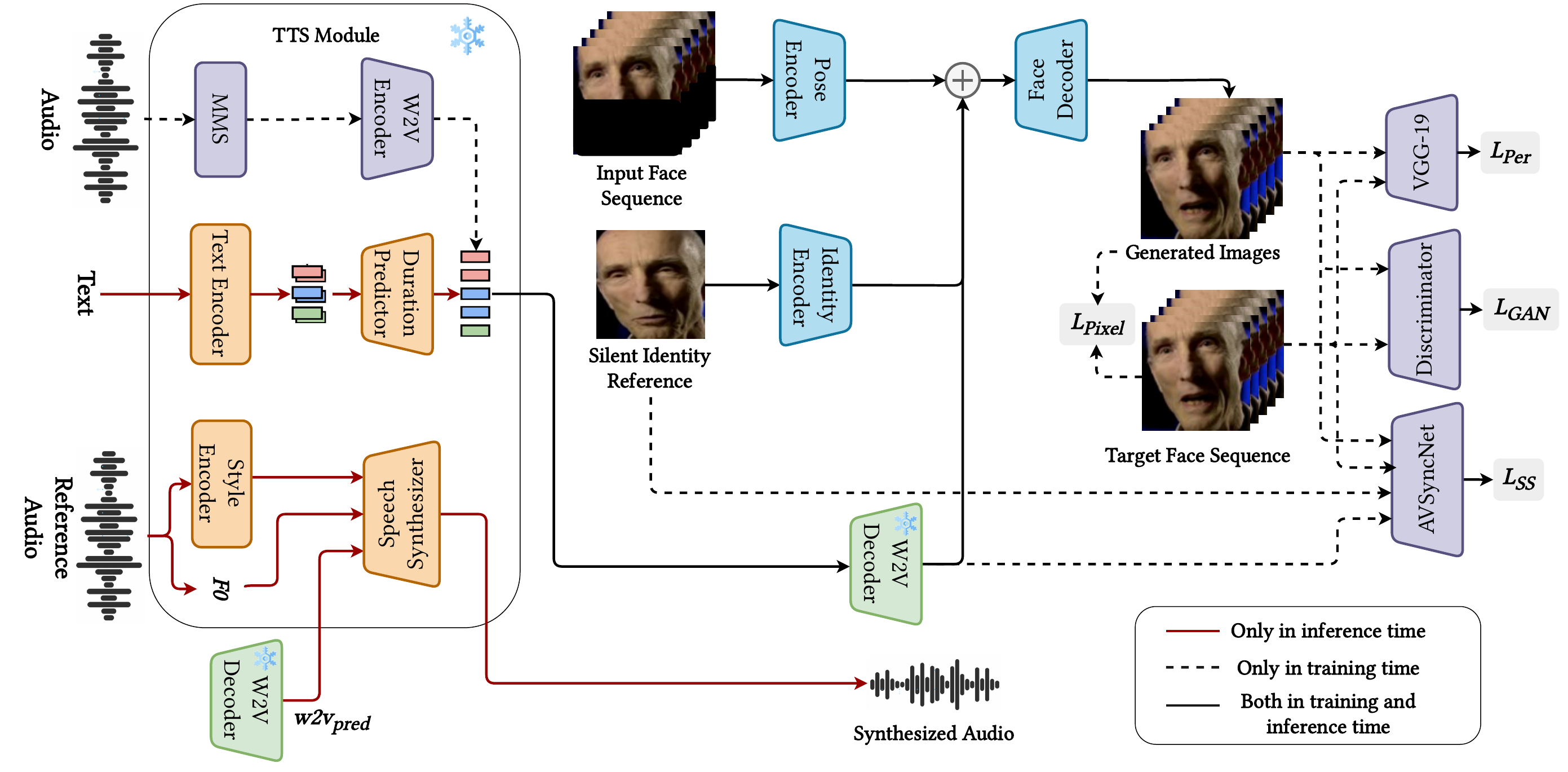}
    \caption{Joint text-to-audio-visual synthesis framework. Text is converted to latent Wav2Vec2 features via TTV, which condition both speech synthesis and talking-face generation for synchronized output without ground-truth audio.}
    \label{fig:model}
\end{figure*}

In order to adapt the talking face generation model for text-to-speech outputs, we used Hierspeech++~\cite{lee2025hierspeech++} as the TTS backbone where we used the outputs from the text-to-vec module as inputs to the TFG module. Then TFG and speech synthesizer synthesize the speech and corresponding talking face. The overall architecture is shown in Figure~\ref{fig:model}.

\subsection{TTS Module}

HierSpeech++ is a hierarchical speech synthesis model that combines linguistic, acoustic, and prosodic representations to generate natural and expressive speech. Unlike conventional TTS systems that operate on mel-spectrograms, HierSpeech++ leverages hierarchical latent representations derived from the self-supervised speech model Wav2Vec2 
 (W2V2)~\cite{baevski2020wav2vec}, trained on massively multilingual data~\cite{pratap2024scaling}, and aligns them with text through a conditional variational autoencoder architecture. This design enables improved prosody modeling, robustness to out-of-domain text, and enhanced expressiveness.

In our pipeline, we employ TTV and speech synthesizer modules. The TTV module is a variational autoencoder similar to VITS~\cite{kim2021conditional}, trained to synthesize W2V2 embeddings and F0 from text. It consists of a text encoder for generating text embeddings, a W2V2 encoder-decoder for reconstructing W2V2 features, and a duration predictor that learns text-to-W2V2 alignment via monotonic alignment search (MAS). During training, we use the W2V2 encoder-decoder to reconstruct ground-truth W2V2 embeddings and employ the predicted embeddings to fine-tune the model for synthetic speech adaptation, ensuring that video–audio alignment from the original data is preserved, a critical factor for talking-face generation training.

During inference, as ground-truth audio is unavailable, we generate W2V2 features from text using the predicted durations and feed these predicted features into the pipeline to jointly synthesize speech and the corresponding talking face. Reference audio is used for style conditioning, including speaker identity, while the hierarchical speech synthesizer generates the waveform. This approach enables tight synchronization between generated speech and facial motion while maintaining naturalness and speaker characteristics.

\begin{table*}[t]
    \centering
    \footnotesize
    \begin{tabular}{l|cccccc|cccccc}
    \toprule
    & \multicolumn{6}{c|}{With Real Data} & \multicolumn{6}{c}{With TTS Data} \\
    \midrule
        Method & SSIM$\uparrow$ & PSNR$\uparrow$ & FID$\downarrow$ & LSE-C$\uparrow$ & LSE-D$\downarrow$ & CSIM$\uparrow$ & SSIM$\uparrow$ & PSNR$\uparrow$ & FID$\downarrow$ & LSE-C$\uparrow$ & LSE-D$\downarrow$ & CSIM$\uparrow$ \\
    \midrule
         Wav2Lip~\cite{prajwal2020lip} & 0.86 & 26.53 & 7.05 & 7.59 & 6.75 & 0.84 & 0.94 & 30.71 &  10.85   & 6.18  & 8.12  & 0.86 \\
         TalkLip~\cite{wang2023seeing} & 0.86 & 26.11  & 4.94 & {8.53} & {6.08} & 0.75 & 0.84 & 24.33 &  12.81   & 7.05  & 7.21  & 0.76 \\
         IPLAP~\cite{zhong2023identity} & {0.87} & {29.67} & {4.10} & 6.49 & 7.16 & 0.82 & 0.88 & 28.27 & 11.47    & 6.51  & 7.08  & 0.83  \\
         AVTFG~\cite{yaman2024audio} & 0.95 & 31.27 & 4.51 & 7.95 & 6.30 & 0.80 & 0.94 & 32.98 &  13.67   & 6.19  & 8.16  & 0.88 \\
         PLGAN~\cite{yaman2024audiodriventalkingfacegeneration} & {0.95} & {32.64} & {3.83} & {8.41} & {6.03} & 0.79 & 0.94 & 31.21 & 12.01   & 6.30  & 7.99  & 0.87 \\
         Diff2Lip~\cite{mukhopadhyay2024diff2lip} & 0.94 & 31.68 & 3.80 & 7.87 & 6.46 & 0.85 & 0.93 & 30.61 & 15.37  & 7.06  & 6.84  & 0.87 \\
         \midrule
         Ours & 0.92 & 30.90 & 4.93 & 7.97 & 6.18 & 0.85 & 0.93 & 31.48 & 13.15 & 6.30 & 7.81 & 0.84 \\
    \bottomrule
    \end{tabular}
    \caption{Quantitative results of our talking face generation model compared with SOTA methods. The left part shows generation with real audio (and Wav2Vec2 features extracted from real audio for our model), while the right part shows generation with TTS-generated audio (and TTS-predicted features for our model).}
    \label{tab:TFG_quantitative_results}
\end{table*}

\begin{figure*}[t]
    \centering
    \includegraphics[width=1\linewidth]{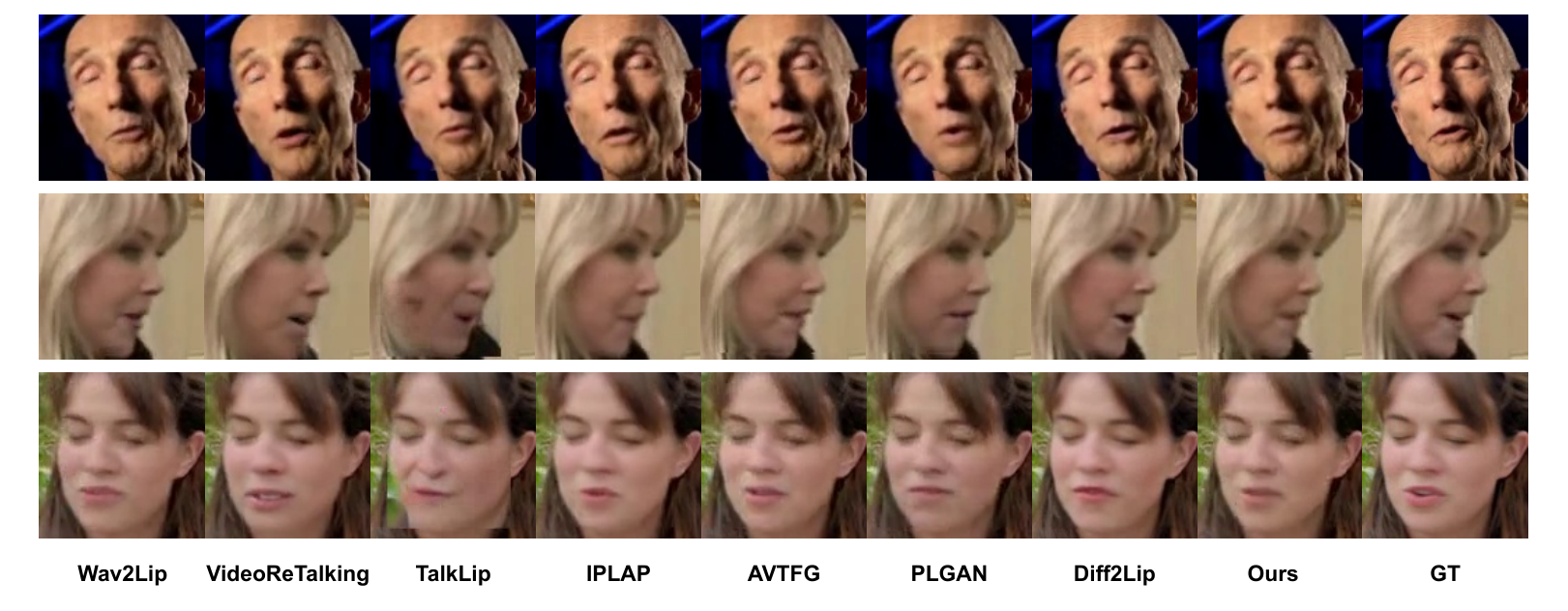}
    \vspace{-0.1cm}
    \caption{Qualitative comparison of our model with other approaches. Note that since our model is trained with predicted Wav2Vec2 features and is designed to align lips with TTS-generated audio in a joint space, the expected lip shapes do not necessarily need to match those of the GT.}
    \label{fig:qualitative_results}
\end{figure*}

\subsection{Talking Face Generation Model}
\label{sec:talking_face_generation}
We use the GAN-based~\cite{goodfellow2014generative} talking face generation model presented in \cite{yaman2024audiodriventalkingfacegeneration}.
The original model includes two image encoders responsible for processing the identity reference image and the input face sequence to generate embeddings, as well as an audio encoder that extracts audio embeddings from mel-spectrogram input. 
However, since our goal is to generate video from the same latent space as TTS, we remove the audio encoder from the architecture. 
Instead, we utilize features from the joint space and directly concatenate them with the visual embeddings. 
Finally, the face decoder generates the face sequence with synchronized lip movements.
In \cite{yaman2024audio}, it was also observed that the identity reference can occasionally harm training stability and the model’s lip-sync performance due to the lip leaking problem. 
To address this, the authors proposed using an additional preprocessing network to modify the selected identity reference image and generate a silent-face image, representing a face with a stable, closed mouth. 
Since this approach improves both performance and training stability, we adopt the same strategy in training our model.

\noindent\textbf{Two-stage training strategy.}
We propose a two-stage training strategy for our talking-face generation model to ensure tight synchronization with TTS-generated speech.
In the first stage, we extract audio features from pretrained W2V2 model~\footnote{https://huggingface.co/facebook/mms-300m}
 that match the output space of the TTV module from HierSpeech++. These features are used as audio conditions to train the model, providing a robust initial mapping from speech representations to facial motion.

In the second stage, we finetune the model using features predicted by the TTV decoder of the TTS model. This step is crucial for adapting to the distribution shift between clean, pretrained W2V2 features and the synthetic TTS-predicted vectors, which may have different statistics. Unlike traditional cascaded pipelines, raw audio is unavailable during inference; the face generator must rely on the same predicted features used for speech synthesis.

\begin{table*}[h!]
    \centering
    \footnotesize
    \begin{tabular}{l|cccccc|cccccccc}
    \toprule
    & \multicolumn{6}{c|}{With Real Data} & \multicolumn{6}{c}{With TTS Data} \\
    \midrule
    Method & SSIM$\uparrow$ & PSNR$\uparrow$ & FID$\downarrow$ & LSE-C$\uparrow$ & LSE-D$\downarrow$ & CSIM$\uparrow$ & SSIM$\uparrow$ & PSNR$\uparrow$ & FID$\downarrow$& LSE-C$\uparrow$ & LSE-D$\downarrow$ & CSIM$\uparrow$ \\
    \midrule
    Wav2Lip & 0.84 & 25.84 & 7.89 & {7.35} & 7.18 & 0.74 & 0.87 & 23.21  & 11.93 & 6.82  & 7.49  & 0.76 \\
    TalkLip & 0.85 & 25.70 & 4.04 & 6.04 & 8.21 & 0.74 & 0.86 & 24.65  & 12.64 & 4.41   & 9.54   & 0.74 \\
    IPLAP & 0.86 & {28.98} & 3.95 & 3.63 & 10.10 & 0.77 & 0.91 & 29.74  & 11.25 & 3.49  & 10.50  & 0.81 \\
    AVTFG & 0.85 & 26.43 & 5.78 & 6.84 & 7.90 & 0.72 & 0.89 & 26.99  & 16.07 & 5.83  & 8.58  & 0.73 \\
    PLGAN & 0.86 & 25.38 & 4.11 & 7.58 & 6.81 & 0.73 & 0.87 & 25.57  & 15.99    & 6.91  & 7.22   & 0.74 \\
    Diff2Lip & 0.92 & 30.32 & 3.59 & 6.71 & 7.26 & 0.83 & 0.93  & 31.11 & 8.05 & 6.21   & 7.36  & 0.85 \\
    \midrule
    Ours & 0.92 & 31.35 & 3.97 & 4.39 & 8.72 & 0.83 & 0.93 & 31.37 & 15.06 & 5.38 & 8.68 & 0.84 \\
    \bottomrule
    \end{tabular}
    \caption{Cross-test evaluation of the models. Instead of using matching audio–video pairs, we randomly pair audio and video to create a mismatched test setup. As in Table \ref{tab:TFG_quantitative_results}, the left and right parts correspond to generation with real and TTS-generated data, respectively.}
    \label{tab:cross_test}
\end{table*}

\noindent\textbf{Loss functions.}
In training our model, we employ the following loss functions:
(1) Adversarial loss~\cite{goodfellow2014generative}: A discriminator network is used to compute adversarial loss based on its output, guiding the model toward generating realistic outputs.
(2) Perceptual loss~\cite{johnson2016perceptual}: We adopt a pretrained VGG-19 model~\cite{simonyan2014very} to extract features from both the generated and GT faces, and compute the L2 distance between them. 
This loss contributes to visual quality and identity preservation.
(3) Pixel reconstruction loss: We compute the L1 distance between the generated and GT faces in pixel space, which helps preserve fine visual details.
(4) Stabilized synchronization loss: Following \cite{yaman2024audio}, we use the stabilized synchronization loss, which outperforms vanilla lip-sync loss~\cite{prajwal2020lip} and other lip-sync learning methods. 
However, during the second training stage, since the GT data may not be perfectly aligned with the predicted features, we employ vanilla lip-sync loss instead. 
Given that our model has already learned lip synchronization in the first stage, we apply vanilla lip-sync with a small coefficient during the second stage of training.

\section{Experimental Results}
\label{sec:experimental_results}


\noindent\textbf{Evaluation.}
We train our talking face generation model on the LRS2 training set and evaluate it on the LRS2 test set.
For evaluation of talking face generation module, we follow the standard setup in the literature and employ widely used metrics. 
To assess visual quality, we report SSIM~\cite{wang2004image}, PSNR, and FID~\cite{heusel2017gans}. 
For audio–lip synchronization, we use mouth landmark distance (LMD)~\cite{chen2019hierarchical} and LSE-C \& LSE-D~\cite{chung2017out,prajwal2020lip}. 
LMD measures the distance between the mouth landmarks of the generated and GT faces, while LSE-C and LSE-D rely on the SyncNet model~\cite{chung2017out} to extract audio–visual features and compute confidence and distance, respectively.

For evaluating the speech synthesizer on LRS2 dataset, we measure the word error rate (WER) using Whisper Large-v3~\footnote{https://huggingface.co/openai/whisper-large-v3} for intelligibility assessment, speaker embedding cosine similarity (SECS) using Resemblyzer~\footnote{https://github.com/resemble-ai/Resemblyzer} for speaker identity preservation assessments and UTMOS~\cite{saeki2022utmos} for perceived naturalness. 


\subsection{Results}
\noindent{\textbf{Talking face generation.}} Table \ref{tab:TFG_quantitative_results} reports the quantitative results on the LRS2 test set mactched audio-video scenario, comparing our model with existing methods. 
We consider two setups in this experiment. 
On the left, videos are generated with the compared models using real audio, while our model uses Wav2Vec2 features extracted from the same audio. 
On the right, audio is generated with our TTS model and provided to the compared models, whereas our model uses the corresponding TTS-predicted Wav2Vec2 features.
In terms of visual quality, our model achieves nearly state-of-the-art (SOTA) performance on SSIM, PSNR, and FID.
For identity preservation, measured by CSIM, we obtain the best score together with Diff2Lip.
For lip synchronization, our model outperforms most of the methods.



We further conduct a cross-test evaluation to assess the models under more challenging conditions, where audio and video are randomly paired, in contrast to the matched (GT) pairs used in Table \ref{tab:TFG_quantitative_results}. 
The results are presented in Table \ref{tab:cross_test}. 
For identity preservation, our model achieves SOTA performance, consistent with the matched scenario. 
In terms of FID, our score is slightly below SOTA. 
For lip synchronization, our model demonstrates moderate performance according to the LSE-C and LSE-D metrics. 
Note that the setup in this table follows the same protocol as Table \ref{tab:TFG_quantitative_results} with respect to real and TTS data; the only difference is the use of mismatched audio–video pairs.



\begin{table}[]
    \centering
    \resizebox{\linewidth}{!}{
    \begin{tabular}{l|cccccc}
    \toprule
    Method & SSIM$\uparrow$ & PSNR$\uparrow$ & FID$\downarrow$ & LSE-C$\uparrow$ & LSE-D$\downarrow$ & CSIM$\uparrow$ \\
    \midrule
    Ours - first stage only w/ Real  & 0.91 & 29.78 & 7.29 & 8.39  & 5.92 & 0.84  \\
    Ours - first stage only & 0.92 & 30.42 & 9.09 & 3.24 & 11.12 & 0.84 \\
    Ours - two-stage, no sync   & 0.92 & 31.21 & 7.47 & 3.31 & 10.81 & 0.84 \\
    Ours - full                 & 0.93 & 31.48 & 5.31 & 4.14 & 10.28 & 0.84 \\ 
    \bottomrule
    \end{tabular}}
    \caption{Ablation study evaluating the impact of the proposed training strategy.}
    \label{tab:ablation}
\end{table}
\noindent{\textbf{Speech Synthesis.}} 
The evaluation results of HierSpeech++ outputs compared to ground-truth speech are reported in Table~\ref{tab:tts_eval}. The WER scores indicate that HierSpeech++ generates highly intelligible speech from text, even surpassing the performance on the ground-truth recordings, likely due to the dataset containing suboptimal recording conditions, whereas the TTS output is cleaner and less noisy. SECS results show that speaker identity is largely preserved, and UTMOS scores suggest that the synthesized speech maintains naturalness comparable to real speech.
\begin{table}[t!]
\centering
\small
\begin{tabular}{lccc}
\hline
Method & WER$\downarrow$ & SECS$\uparrow$ & UTMOS$\uparrow$ \\
\hline
GT & 4.47\% & - & 3.05 \\
HierSpeech++ & 1.51\% & 72\% & 4.22 \\
\hline
\end{tabular}
\caption{Quantitative evaluation of the synthesized speech.}
\label{tab:tts_eval}
\end{table}


\subsection{Ablation Study}
\label{sec:ablation_study}

Table \ref{tab:ablation} presents the ablation study of our method. 
We first evaluate the model trained only in the first stage, using Wav2Vec2 features extracted from real audio (\textit{Ours – first stage only w/ Real}). 
Next, we apply the same model with TTS-predicted features, reported as \textit{Ours – first stage only}. 
We then evaluate a two-stage model in which the second stage is trained without any explicit lip-sync loss (\textit{Ours – two-stage, no sync.}). 
Finally, the last row corresponds to our full pipeline.

\section{Conclusion}
We present a joint text-to-audio-visual synthesis framework using latent speech representations from HierSpeech++. By conditioning the talking-face generator on TTS-predicted Wav2Vec2 features, we achieve tight audio–visual alignment, preserve speaker identity, and generate natural speech with synchronized facial motion, compatible to other models. Limitations include reliance on high-quality latent features, which may reduce generalization to unseen languages or noisy TTS outputs, and the lack of explicit modeling for subtle facial expressions beyond lip movements.


%
%
%


\vfill\pagebreak


\bibliographystyle{IEEEbib}
\bibliography{strings,refs}

\end{document}